\documentclass[runningheads]{llncs}
\usepackage{graphicx}
\usepackage{xcolor}
\usepackage[linesnumbered,ruled,vlined]{algorithm2e}
\usepackage{bm}
\usepackage{microtype}
\usepackage{booktabs}
\usepackage{graphicx}
\usepackage{epsfig}
\usepackage{epstopdf}
\usepackage{amsfonts, amsmath, amssymb, color, mathrsfs, cite}   

\usepackage{array, boldline, makecell, booktabs}

\usepackage{enumitem}
\setlist[itemize]{noitemsep, topsep=0pt}
\setlist[enumerate]{noitemsep, topsep=0pt}

\DeclareMathOperator*{\argmax}{arg\,max}
\DeclareMathOperator*{\argmin}{arg\,min}

\newcommand{\Osymbol}{{\mathcal O}}
\newcommand{\BO}[1]{\Osymbol\left(#1\right)}

\newcommand{\E}[1]{\textrm{\bf E}\left[#1\right]}
\renewcommand{\Pr}[1]{\textrm{\bf Pr}\left[#1\right]}

\newcommand{\mA}{\mathcal{A}}

\newcommand{\mH} {\mathcal{H}}
\newcommand{\mU} {\mathcal{U}}
\newcommand{\mO} {\mathcal{O}}

\newcommand{\mX} {\mathbb{X}}
\newcommand{\mS} {\mathbb{S}}

\newcommand{\mZ} {\mathbb{Z}}
\newcommand{\mR} {\mathbb{R}}
\newcommand{\Rd} {\mathbb{R}^d}

\newcommand{\bx}{\mathbf{x}}

\newcommand{\bq}{\mathbf{q}}

\newcommand{\ba}{\mathbf{a}}

\newcommand{\ballqr} {B_{\mS}(\bq,r)}
\newcommand{\dotxa}{\bx^\top \ba}

\begin{document}
	
\title{An Efficient Hashing-based Ensemble Method for Collaborative Outlier Detection
}

%
%


\author{Kitty Li \and
	Ninh Pham}
%
%
\institute{School of Computer Science, University of Auckland, New Zealand \\
	\email{dli459@aucklanduni.ac.nz, ninh.pham@auckland.ac.nz}}
\maketitle              

\begin{abstract}
In collaborative outlier detection, multiple participants exchange their local detectors trained on decentralized devices without exchanging their own data.
A key problem of collaborative outlier detection is efficiently aggregating multiple local detectors to form a global detector without breaching the privacy of participants' data and degrading the detection accuracy. 
We study locality-sensitive hashing-based ensemble methods to detect collaborative outliers since they are mergeable and compatible with differentially private mechanisms.
Our proposed LSH iTables is simple and outperforms recent ensemble competitors on centralized and decentralized scenarios over many real-world data sets.

\end{abstract}
\section{Introduction}

We study the unsupervised \emph{collaborative} outlier detection, the task of finding unsupervised outliers in the collaborative environment where multiple participants are willing to share their own trained model and to utilize other's trained models to increase their detection accuracy.
Collaborative outlier detection has many vital applications in medicine,
banking, and finance, where data sources are sensitive and often belong to several participants. 
For instance, on medical screenings, it is challenging for hospitals or clinics to provide a highly accurate diagnosis of diseases with respect to a broader population while protecting the patients’ privacy and their medical records.

Besides the traditional demands of efficiency and effectiveness, collaborative outlier detection 
requires solutions that use less communication cost for broadcasting participants' trained models and preserve the privacy of participants' data.
Recent works on collaborative outlier detection ensure the data privacy via Secure Multi-Party Computing (MPC) protocols~\cite{PLOF,PIF2,LDOF} that often suffer a high communication
and computational overhead for model aggregation.

We study another direction that asks to release local trained models under differential privacy~\cite{DP}, a de facto standard for statistical computations over databases that contain private data.
This approach enables individual participants to aggregate other's differentially private released models to form a global model and perform detection on their local data.
Our decentralized approach that executes outlier detection on local devices is different from recent differential privacy solutions~\cite{DP15,PCOR} that aim at detecting outliers on centralized data.

Isolation tree-based~\cite{iForest, LSHiForest} and proximity-based~\cite{EnskNN} unsupervised ensemble methods have shown their efficiency and accuracy on the centralized scenario.
However, without sharing the data, their performance has deteriorated when evaluating on small local data since hidden outliers only exhibit their outlier behaviour on the wider population.
This work studies unsupervised hashing-based ensemble methods~\cite{RS,ACE} to detect collaborative outliers due to their \emph{mergeability}~\cite{Mergeable}.
The mergeable property is that we can merge two histograms of two data sets into a single histogram of the union of these two data sets (via algebraic operators such as sum or max) while preserving the property of these histograms.
Therefore, we can quickly compute the global histogram of the union of local data and execute outlier detection based on this global histogram on local devices.


In particular, we study and generalize the recent randomized subspace hashing (RS-H) ensemble~\cite{RS} to detect unsupervised outliers in high dimensions.
We show that RS-H on rotated data is essential  locality-sensitive hashing (LSH)~\cite{LSH}. 
We propose \emph{LSH iTables}, an LSH based ensemble using a random feature-based LSH family that requires $O(1)$ hash evaluation time.
We establish a connection between LSH iTables and the conventional distance-based outlier detection~\cite{rNN}.
Since the histograms derived from LSH tables are mergeable (by a simple sum operation) and require small space storage (e.g. a few KB), each participant can efficiently broadcast their constructed LSH-based histograms and aggregate the others' histograms to build a global model.
To enhance LSH iTables with differential privacy, we add a sufficient level of Laplacian noise~\cite{DPBook} into the LSH-based histogram before releasing it so that it is impossible to infer the presence or absence of any single instance of local data from the global model.

Empirically, LSH iTables provides competitive detection accuracy and runs significantly faster than many advanced ensemble methods on centralized data.
On the decentralized setting, differentially private LSH iTables with small privacy loss budgets achieve similar accuracy as its non-privacy version.

\section{Preliminary}

\subsection{Locality-sensitive hashing}

Locality-sensitive hashing (LSH)~\cite{LSH} is an algorithmic primitive for the approximate nearest search problem in high dimensions with sublinear query time guarantees. 
In principle, LSH hashes near points into the same bucket with high probability, while far away points tend to be hashed into different buckets. 
The LSH definition is as follows.

\begin{definition}
Fix a distance function $d(\cdot, \cdot): \Rd \times \Rd \in \mathbb{R}^+$. 
For positive real values $r_1, r_2, p_1, p_2$ and $r_2 > r_1, p_1 > p_2$, a family $\mH$ is called $(r_1, r_2, p_1, p_2)$-sensitive if for a uniformly chosen $h \in \mH$ and any two points $\bx, \bq \in \Rd$, 
\begin{itemize}
    \item If $d(\bx, \bq) \leq r_1$, then $\Pr{h(\bx) = h(\bq)} \geq p_1$;
    \item If $d(\bx, \bq) \geq r_2$, then $\Pr{h(\bx) = h(\bq)} \leq p_2$.
\end{itemize}
\end{definition}


Since we will show RS-H on rotated data is essentially locality-sensitive, we present the popular random projection-based LSH families (RP-LSH)~\cite{RPLSH} for $\ell_p$ distance based on the $p$-stable distribution properties.

\begin{definition}\label{def:RPLSH}
Given a random vector $\ba$ whose coordinates are randomly selected from the $p$-stable distribution and a scalar $b \sim \mathcal{U}[0, w]$ for a given interval length $w$.
The $p$-stable LSH function for $\ell_p$ distances is computed as $h_{\ba, b}(\bx)=\lfloor \frac{\dotxa + b}{w}\rfloor$.
The random vector $\ba$ is generated from the 1-stable standard Cauchy distribution and the 2-stable normal distribution $N(0, 1)$ for $\ell_1$ and $\ell_2$ distances, respectively.
\end{definition}


Since outliers tend to be far away from the other points, they tend to have fewer points colliding in their buckets.
Therefore, several LSH-based approaches~\cite{ACE,LODA,RS} use the bucket size of a point as its outlier score.


\textbf{An alternative view of the splitting mechanism of iForest.}
iForest~\cite{iForest} is a combination of several binary isolation trees; each is built over a subsample of data.
Each tree partitions a node via a random dimension $i$ and a random splitter value $c_i \sim \mathcal{U}[min_{i}, max_{i}]$ where $min_{i}$ and $max_{i}$ are the minimum and maximum values of all points in that node over dimension $i$.
We will show that this splitting mechanism is essentially an LSH for a specific weighted $\ell_1$ distance.
\begin{definition}\label{def:RFLSH}
Given a data set $\mX$, a random dimension $i \in [d]$ whose values are in range $[min_{i}, max_{i}]$, and a random value $c_i \sim \mathcal{U}[min_{i}, max_{i}]$, we define a hash function $	h_{i, c_i}(\bx)= 1 \text{ if } x_i \geq c_i; \text{ otherwise } 0$.
\end{definition}

It is easy to verify that for $\bx, \bq \in \mX$,
\begin{equation}\label{eq:RF}
	\Pr{h(\bx) = h(\bq)} = 1 - \frac{1}{d} \sum_{i = 1}^{d} \frac{|x_i - q_i|}{max_{i} - min_{i}} \approx e^{-\ell'_1(\bx, \bq) / d} ,
\end{equation}
where $\ell'_1$ is a weighted $\ell_1$ distance and the weight of dimension $i$ is $1 /\left(max_{i} - min_{i}\right)$.
We call this LSH family as \emph{random feature LSH} (RF-LSH) to distinguish with RP-LSH.
While RF-LSH works on $\ell_1$ variants, it requires significantly smaller evaluation time, i.e. $\BO{1}$, compared to $\BO{d}$ of RP-LSH on Definition~\ref{def:RPLSH}.

\subsection{Differential privacy}

Differential privacy (DP)~\cite{DP} is a de facto privacy standard that provides upper bounds for the risk of inferring absence or presence of a single instance in a private database after releasing the analytic model.
Let $\epsilon > 0$ be the privacy loss budget, a randomized $\epsilon$-DP mechanism is defined as follows.

\begin{definition}
Let $\mathcal{A}: \mathcal{X} \mapsto \mO$ be a randomized algorithm that takes any input in $\mathcal{X}$. 
$\mA$ is said $\epsilon$-DP if for all data sets $\mX, \mX' \in \mathcal{X}$ that differ on a single instance and for all possible outputs $\mO$ of $\mA$, we have
$ \Pr{\mA(\mX) \in \mO} \leq e^{\epsilon} \, \Pr{\mA(\mX') \in \mO}$.
\end{definition}

The privacy loss $\epsilon$ is often set to be small to ensure that $\mA$ will output the same result for very similar data sets $\mX$ and $\mX'$  with high probability.
We consider the task of releasing an LSH-based histogram under $\epsilon$-DP (i.e. histogram queries) where the addition or removal of a single instance can affect the
counter in exactly one cell.
Therefore, we study the Laplacian mechanism~\cite{DPBook} that adds Laplacian noise $\eta \sim \text{Lap}(1/\epsilon)$ into each cell to ensure an $\epsilon$-DP released histogram.

\begin{definition}
Given any function $f: \mathcal{X} \mapsto \mR^m$ that constructs an LSH-based histogram of size $m$ for any $\mX \in \mathcal{X}$, the Laplace mechanism is defined as:
$\mathcal{M}_L\left(\mX, f(.), \epsilon \right) = f(\mX) + \left(Y_1, \ldots, Y_m \right)$,
where $Y_i$ are i.i.d random variables drawn from Laplacian distribution $Lap(1/\epsilon)$.
\end{definition}

Due to the composability and post-processing~\cite{DPBook}, the global histogram formed by aggregating $n$ different $\epsilon_i$-DP histograms is also $\sum_{i=1}^n \epsilon_i$-DP and any analytic outcome over the global histogram is also $\sum_{i=1}^n \epsilon_i$-DP. 

\section{LSH iTables for collaborative outlier detection}

For notation simplicity, we use standard notations that $[d] = \{1, \ldots, d \}$. 
Each ensemble method builds $m$ base models over random  subsamples $\mS$ of size $s$ from the data set $\mX$ of size $n$ in $d$ dimensions.
Given an LSH family $\mH$, we concatenate $l$ different hash functions $h_i \in \mH$ to form a new LSH function $g = (h_1, \ldots, h_l)$ to build a hash table.
We denote by $Counter[g(\bq)]$ the bucket size of the testing point $\bq \in \mX$ when using $g$ to construct the table.
For simplicity, we only present the algorithm and its analysis on a base model.

\subsection{A warmup: Rotated randomized subspace hashing}

Sathe and Aggarwal~\cite{RS} propose Randomized Subspace Hashing (RS-H) that constructs hash tables over subsamples on randomly selected subspaces. 
In particular, each base model of RS-H first subsamples a subset $\mS$ of $s = min(1000, n)$ points, then randomly selects $l$ dimensions from $[d]$ to construct a hash table for $\mS$.
Since outliers/inliers tend to distribute on sparse/dense buckets, respectively, the size of the bucket where the point is hashed into is used as an outlier score.

A base model of RS-H needs many internal parameters to control the hash table construction, including (1) the scaling $f \sim \mU[\frac{1}{\sqrt{s}}, 1 - \frac{1}{\sqrt{s}}]$, (2) the number of dimensions $l \sim \mU[1 + 0.5 \log_{\max{(2, 1/f)}}{(s)}, \log_{\max{(2, 1/f)}}{(s)}] = \BO{\log{(s)}}$, and (3) the shifting $\alpha_i \sim \mU[0, f]$ for $i \in [d]$.
For each dimension $i \in [d]$, let $max_i$ and $min_i$ be the maximum and minimum values over $\mS$.
Each base detector hashes the subsampled point $\bx = (x_1, \ldots, x_d) \in \mS$ on the subspace of $l$ selected dimensions.
In particular, it uses $l$ hash functions, each for a selected dimension $i$, that maps $x_i \in \mR \mapsto \bar{x}_i \in \mZ$ where
\begin{align*}
	\bar{x}_i = \Bigl\lfloor \frac{(x_i - min_i)/(max_i - min_i) + \alpha_i}{f} \Bigr\rfloor = \Bigl\lfloor \frac{(x_i - min_i) + \alpha_i(max_i - min_i)}{f(max_i - min_i)} \Bigr\rfloor \, .
\end{align*}

Consider a rotation of $\mS$ where we map any $\bx \in \mS$ to $\bx' = (\bx ^\top \ba_1, \ldots, \bx ^\top \ba_d)$ using $d$ random vector $\ba_i$ whose coordinates are from $N(0, 1)$.
Denote by $max'_i = \argmax_{\bx \in \mS} \bx ^\top \ba_i$, $min'_i = \argmin_{\bx \in \mS} \bx ^\top \ba_i$, the mapping of RS-H on $\bx'$ using the same parameters $f, l, \alpha_i$ on dimension $i$ is $x'_i = \bx ^\top \ba_i \in \mR \mapsto \bar{x}'_i \in \mZ $ where
\begin{align*}
	\bar{x}'_i = \Bigl\lfloor \frac{(\bx ^\top \ba_i - min'_i) + \alpha_i(max'_i - min'_i)}{f(max'_i - min'_i)} \Bigr\rfloor \, . 
\end{align*}

Let $w_i = f(max'_i - min'_i), b_i = \alpha_i(max'_i - min'_i)$.
Since $\alpha_i \sim \mU[0, f]$, $b_i = \alpha_i(max'_i - min'_i) \sim \mU[0, w_i]$. 
The RS-H function at dimension $i$ over the random rotation of $\bx$ is $ x_i \mapsto \bar{x}'_i = \Bigl\lfloor \frac{\bx ^\top \ba_i  + b_i - min'_i}{w_i} \Bigr\rfloor$, which is essential an RP-LSH form.

Since shifting all projected values at dimension $i$ by  $min'_i$ does not change the collision probability, applying RS-H over the random rotation of $\mS$ is identical to using RP-LSH (with different $w$) to construct a hash table on $\mS$. 
The number of dimensions used in RS-H is similar to the number of concatenating LSH functions, and the setting $l = \BO{\log{(s)}}$ above is identical to the standard LSH's setting~\cite{LSH} that discards the contributions of all far away points to the bucket of $\bx$.

Note that the outlier score of a point $\bq$ is proportional to the bucket size where $\bq$ is hashed into. 
The problem of estimating the bucket size is essentially the task of frequency estimation in data stream where the occurrences of $\bq$ (i.e. the bucket size of $\bq$) correspond to the number of points $\bx \in \mS$ that collides with $\bq$.
Therefore, RS-H applies the CountMin sketch~\cite{CountMin} while constructing the hash table to reduce the space usage of RS-H for streaming data.

\subsection{LSH iTables}

Leveraging the observation of rotated RS-H, we propose \emph{LSH iTables}, an LSH-based ensemble method using the RF-LSH family for unsupervised outlier detection due to its $\BO{1}$ hash evaluation time.
The base model constructs a hash table using $l$ random RF-LSH functions on a subsample $\mS$ of size $s$.
It maintains a histogram $Counter$ of size $2^l$ that contains the number of points on each bucket.
Before releasing this histogram, we add a Laplacian noise Lap$(1/\epsilon)$ where $\epsilon$ is the privacy budget into each cell to preserve the differential privacy.
Algorithm~\ref{alg:train} shows how to construct the local histogram on local data.

\begin{algorithm}[!t]
	\SetKwInput{KwInput}{Input}                
	\SetKwInput{KwOutput}{Output}              
	\KwInput{A subsample $\mS$ of size $s$ on local data, the shared LSH functions $g = (h_1, \ldots, h_l)$}
	\KwOutput{An array $Counter$ of size $2^l$}
	Initialize all entries of $Counter$ as 0
	
	\For{ each point $\bx \in \mS$}
	{
		$g(\bx) \gets (h_1(\bx), h_2(\bx), \ldots, h_l(\bx)) \in \{0, 1\}^l$
		
		$Counter[g(\bx)] = Counter[g(\bx)] + 1$ 
	} 	
	\For{ each $i \in [2^l]$ }
	{
		$Counter[i] = Counter[i] + \text{Lap}(1/\epsilon)$ \tcp*{Add noise to preserve $\epsilon$-DP}  
	}
	Release $Counter$ 
	\caption{Training phase of a base model on local data}
	\label{alg:train}
\end{algorithm}
\begin{algorithm}[!t]
	\SetKwInput{KwInput}{Input}                
	\SetKwInput{KwOutput}{Output}              
	
	\KwInput{Local data $\mX$; the shared LSH functions $g = (h_1, \ldots, h_l)$,  histograms of other participants}
	\KwOutput{the list $L$ of outlier scores for $\mX$}
	Aggregate histograms of other participants to achieve the global $Counter$.
	
	\For{each point $\bq$  $\in \mX$}
	{  
		$g(\bq) \gets (h_1(\bq), h_2(\bq), \ldots, h_l(\bq)) \in \{0, 1\}^l$
		
		$score \gets \log_2{\left( \max(Counter[g(\bq)], 1 \right)}$  \tcp* {to handle negative value due to the Laplacian noise} 
		
		Add $score$ as the outlier score of $\bq$ into the list $L$ 
	}
	return $L$
	\caption{Testing phase of a base detector on local data}
	\label{alg:test}
\end{algorithm}

Given that all participants construct the base model using the same LSH functions,\footnote{We can use a trust party or MPC protocols to hide the hashing mechanism, e.g.~\cite{PIF2}.} these LSH-based histograms are seamlessly mergeable.
After merging other's histograms to form the global histogram over all participants' data, the outlier score of $\bq$ computed based on its bucket size, as shown in Algorithm~\ref{alg:test}.
Similar to other ensemble methods~\cite{Bias}, LSH iTables takes the average of outlier scores provided by $m$ base detectors for outlier ranking.
We note that the outlier detection is executed on the local device.
Hence, given $\epsilon$-DP released histograms of the base models, the outputs of LSH iTables are differentially private.

\subsection{Theoretical analysis}

We start from a conventional distance-based outlier definition and show how the subsampling mechanism can preserve the distance-based outlier score.
After that, we describe how LSH can speed up the outlier score computation process.
All analysis is on local data and can be generalized on global data due the mergeablity of the LSH iTables histogram.
Due to the lack of space, we skip the analysis of adding  Laplacian noise Lap$(1/\epsilon)$ to ensure $\epsilon$-DP.

\textbf{Subsampling.} Recall the distance-based outlier concept~\cite{rNN}, given a fix radius $r$ and a distance function $\ell_1'(\cdot, \cdot)$, we compute the set $B(\bq, r) = \{\bx \in \mX \, | \, \ell_1' (\bq,\bx) \leq r\}$ for any testing point $\bq \in \mX$. 
For outlier ranking, we use $|B(\bq,r)|$ as an outlier score, and the smaller $|B(\bq,r)|$ is, the more likely $\bq$ is an outlier.
Given a random subsample set $\mS$ of size $s$, we denote by $|B_{\mS}(\bq, r)|$ the number of points in $\mS$ within the radius $r$ to $\bq$.
Since any $\bx \in \mX$ is sampled with the same probability $s/n$, we have
	$\E{|B_{\mS}(\bq, r)|} = (s/n) |B(\bq, r)|  \, .$	


\textbf{Speeding up with LSH.} We observe that estimating $|B_{\mS}(\bq,r)|$ can be solved faster by utilizing LSH to avoid the significant pairwise distance computations.
We combine $l$ independent RF-LSH functions from an $(r_1, r_2, p_1, p_2)$-sensitive LSH family $\mH$ to form a new LSH function $g = \{h_1, \ldots, h_l\}$.
For any $\bx \in \mS$, we consider $r_2 = r$ and hence if $\ell_1'(\bx, \bq) \geq r$, then $\Pr{g(\bq) = g(\bx))}\leq {p_2}^ l$.
We set $p_2^l = 1/s^2$ and hence $l = 2\log{(s)} / \log{(1/p_2)} = \BO{\log{(s)}}$ to make sure that the expected contributions of all $\bx \in \mS \setminus \ballqr$ to the bucket $g(\bq)$ is at most $1/s$.
Therefore, we can use the bucket size of $\bq$, i.e. $Counter[g(\bq)]$, as an estimate of $|\ballqr|$. 
%
The following lemma justifies the correctness of LSH iTables.

\begin{lemma}\label{lm:LSH}
Assume that for any $\bx \in \mS$, if $\bx \in \ballqr$ then $\ell_1'(\bx, \bq) \sim \mU[0, r]$.
Consider two testing points $\bq_1$ and $\bq_2$, if $|B_{\mS}(\bq_1, r)| > |B_{\mS}(\bq_2, r)|$, then we have $Counter[g(\bq_1)] \geq Counter[g(\bq_2)]$ in expectation.
\end{lemma}

\begin{proof}
Given any testing point $\bq \in \mX$, for any $\bx \in \mS$, we have
\begin{align*}
	\E{Counter[g(\bq)]} = \sum_{\bx \in B_{\mS}(\bq, r)}{\Pr{g(\bx) = g(\bq)}} + \sum_{\bx \notin B_{\mS}(\bq, r)}{\Pr{g(\bx) = g(\bq)}} \,.
\end{align*}
Let $p(r)$ be the collision probability of RF-LSH at distance $r$.
We have $p(cr) \approx p(r)^c$ for any constant $c > 1$ (see Equation~\ref{eq:RF}).
Assume that if $\bx \in \ballqr$ then $\ell_1'(\bx, \bq) \sim \mU[0, r]$, in expectation, for two testing points $\bq_1, \bq_2$ we have 		
\begin{align*}
\E{Counter[g(\bq_1)]} &= |B_{\mS}(\bq_1, r)| p(r/2) + \sum_{\bx \notin B_{\mS}(\bq_1, r)}{\Pr{g(\bx) = g(\bq_1)}}  \\ 
&\geq |B_{\mS}(\bq_1, r)| p(r/2) \, . \\
\E{Counter[g(\bq_2)]} &= |B_{\mS}(\bq_2, r)| p(r/2) + \sum_{\bx \notin B_{\mS}(\bq_2, r)}{\Pr{g(\bx) = g(\bq_2)}}  \\
&\leq |B_{\mS}(\bq_2, r)| p(r/2) + 1/s \, .
\end{align*}
The last inequality is due to the fact that we use $l = 2\log{(s)} / \log{(1/p_2)}$, the total collision probability of all $\bx \notin B_{\mS}(\bq_2, r)$ is at most $1/s$ in expectation.
Since $|B_{\mS}(\bq_1, r)| > |B_{\mS}(\bq_2, r)|$, $  p(r/2) \approx \sqrt{p(r)} = 1/s$, we prove the claim. \qed
\end{proof}

\textbf{Parameter settings.} We follow the setting suggested by RS-H by setting $s = min(1000, n)$ and selecting number of hash functions randomly, e.g. $l \sim \mU(1 + 0.5\log_{\max{(2, 1/f)}}{(s)}, \log_{\max{(2, 1/f)}}{(s)}) = \BO{\log{(s)}}$ where $f \sim \mU(\frac{1}{\sqrt{s}}, 1 - \frac{1}{\sqrt{s}})$.
We note that different outliers on different density areas require different values of $r$.
Therefore, the random choice of $l$ adds more diversity to the base model and increases the accuracy of LSH iTables.

\textbf{The complexity.} Since we use $l = \BO{\log{(s)}}$, the training time complexity is $O(ms\log{s})$ while the testing time complexity is $O(nm\log{s})$.
Each base model requires a constant space $\BO{s}$ to store the histogram.
This means that it is extremely fast to broadcast the trained model to other participants and merge other trained models to exploit multiple data sources.


\section{Experiment}
We implement LSH iTables and other ensemble competitors, including Ensemble KNNW~\cite{EnskNN}, Ensemble LOF~\cite{EnskNN}, iForest~\cite{iForest}, LSH iForest~\cite{LSHiForest}, RS-H~\cite{RS}, LODA~\cite{LODA}, and ACE~\cite{ACE}, in C++.
We compile with -O3 optimization and conduct experiments on a 2.80 GHz core i5-8400 32GB of RAM with a single CPU.
%
We use the standard AUC score (i.e. area under the ROC curve) to evaluate the accuracy of unsupervised ensemble detectors since they output outlier rankings.
For measuring efficiency, we report the total running time of each detector.
We run all ensembles~10 times to get the average AUC.
Since the standard deviation is very small, we do not report it.

We present empirical evaluations on real-world data sets to verify our claims, including (1) LSH iTables outperforms all studied ensembles in terms of accuracy and efficiency, and (2) LSH iTables is suitable to detect collaborative outliers due to the mergeability and compatibility with $\epsilon$-DP mechanisms.

\textbf{Data sets.}
We use public data sets from ODDS\footnote{http://odds.cs.stonybrook.edu/} (BreastW, Pima, Cardio, Mnist, Musk, Pendigits (Pen), Satimage-2 (Sat), Thyroid, and Shuttle) and UCI\footnote{http://archive.ics.uci.edu/ml/index.php} (KDD99, Covertype), as shown in Table~\ref{tb:dataset}. 
We pre-process them by only removing duplicates if exists.

\begin{table*}[b]
	\centering
	\caption{Data sets: short names, \# points $n$,  \# dimensions $d$, and  \# outliers $o$.}
	\begin{tabular}{|c|c|c|c|c|c|c|c|c|c|c|c|}
		\hline
		& BreastW & Pima & Cardio & Mnist & Musk & Pen & Sat  & Thyroid & Shuttle & Cover & KDD99  \\ \hline
$n$    & 683 & 768 & 1822   & 7603  & 3062    & 6870   & 5801        & 3656     & 49097      & 286048   & 48113     \\ \hline
$d$     & 9 & 8 & 21     & 100   & 166     & 16     &  36         & 6       & 9        & 54     & 40          \\ \hline
$o$     &239 & 268 & 175    & 700   & 97      & 156    & 69          & 93    & 3511      & 2747      & 200        \\ \hline
	\end{tabular}
	\label{tb:dataset}
\end{table*}

\textbf{Parameter settings.} 
We set the number of base detectors $m = 100$ for all ensembles.
Ensemble KNNW (that takes the average of KNN distances) and LOF uses $k = 5$  and $s = min(0.1n, 1000)$ to break the quadratic time complexity of computing all pairwise distance.
LSH iForest and iForest use $s = 256$ with the height limit of 8.
LSH iForest with $\ell_2$ LSH family uses $w = 4$ as suggested in~\cite{LSHiForest}.
ACE uses the number of LSH functions $l = 15$.
Both ACE and LODA use $s = n$. 
RS-H uses $s = min(1000, n)$ with CountMin sketch size $4 \times 1000$.

\subsection{AUC and running time on centralized data}

\begin{table*}[!ht]
	\centering
	\caption{AUC of ensemble methods. The top-2 AUCs are in boldface.}
	\begin{tabular}{|c !{\vline width 1pt} c|c|c|c|c|c|c|c|c|c|c|}
		\hline
		& LSH iTables & RS-H & iForest & LSHiForest & EnsKNNW & EnsLOF  & LODA & ACE \\ \hline
		BreastW     & \textbf{97.3}   & 95.9     & 96.1   & 94.9  & 96.6  & \textbf{97.7}     &  96.6     & 42.6       \\ \hline
		Pima     & \textbf{69.1}   & \textbf{69.0}     & 67.5   & 62.5  & 64.2  & 65.1     & 63.2      & 50.1       \\ \hline
		Cardio     & \textbf{93.4}   & 90.1     & 92.3   & 89.5  & 82.1  & 87.5     & \textbf{93.0}      & 32.4       \\ \hline
		Mnist       & 82.4    & 72.3      & 80.1    & 59.8   & 84.0    & \textbf{87.2}      & 81.6      & \textbf{93.7}         \\ \hline
		Musk       & \textbf{100}    & \textbf{100}      & \textbf{100}    & 51.9   & 84.4    & \textbf{99.2}      & 95.9      & 93.2         \\ \hline
		Pendigits       & \textbf{95.1}    & 90.6      & \textbf{95.4}    & 95.0   & 82.0    & 87.8      & 95.0      & \textbf{95.4}         \\ \hline
		Satimage       & 99.2    & 99.7      & 99.3    & 99.1   & 99.6    & \textbf{99.9}      & 99.3      & \textbf{99.8}         \\ \hline
		Thyroid       &\textbf{94.8}    & 94.5      & \textbf{97.7}    & 90.2   & \textbf{94.8}    & 94.5      & 84.0      & 91.9         \\ \hline
		Shuttle       & 99.0    &\textbf{ 99.2}      & \textbf{99.7}    & 99.0   & 84.1    & 87.0      & 99.3      & 98.9         \\ \hline
		Cover       & \textbf{98.9}    & \textbf{98.3}      & 97.5    & 49.1   & 85.0    & 79.4      & 59.0      & 53.8         \\ \hline
		KDD99       & \textbf{98.8}    & \textbf{99.0}      & 98.7    & 55.8   & 60.7    & 59.2      & 52.8      & 86.3         \\ \hline \hline
		AvgScore & \textbf{93.5}    & 91.7   & \textbf{93.1}     & 77.0   & 83.4  & 85.9    & 83.6   & 76.2 \\ \hline
		AvgRank & \textbf{2.18}    & 3.82   & \textbf{2.55}     & 6.36   & 5.55  & 4.91    & 5.09   & 5.18 \\ \hline
	\end{tabular}
	\label{tb:AUC}
\end{table*}

\begin{table*}[!hb]
	\centering
	\caption{Running time of ensemble methods.}
	\begin{tabular}{|c|c|c|c|c|c|c|c|c|c|c|c|}
		\hline
		& LSH iTables & RS-H & iForest & LSHiForest & EnsKNNW & EnsLOF  & LODA & ACE  \\ \hline
		Pendigits       & \textbf{0.6s}    & 1.2s      & 0.7s    & 7.6s   & 2min    & 2.2min      & 1s      & 1.9s         \\ \hline
		Mnist       & 1s    & 1.8s      & 1.1s    & 8.7s   & 16min    & 17min      & \textbf{0.8s}      & 7.3s         \\ \hline
		Shuttle       & 4.1s    & 7.1s      & 4.5s    & 1.2min   & 15 min    & 15 min      & \textbf{2.4s}      & 10.9s         \\ \hline
		Cover       & \textbf{29s}    & 48s      & 31s    & 4.4min   & 9 hour    & 9 hour      & 50s      & 2.8min         \\ \hline
		KDD99       & \textbf{4.5s}    & 7.8s      & 5.0s    & 48s   & 1.1 hour    & 1.1 hour      & 1.2 hour      & 23s         \\ \hline
	\end{tabular}
	\label{tb:Time}
\end{table*}


This subsection compares the AUC score and running time between LSH iTables and other ensemble methods over~11 data sets.
Table~\ref{tb:AUC} shows that LSH iTables provide the highest average AUC scores among all ensembles.
It also has the highest average ranking over all used real-world data sets.

\textbf{Compare with RS-H.} While LSH iTables and RS-H use a similar random feature-based hashing mechanism, LSH iTables returns higher AUC scores on the majority of data sets.
The gap is significant on Mnist and Pendigits with 10.1\% and 4.5\%, respectively.
Regarding the average rank, LSH iTables is top-1 while RS-H is top-3.
Consider $g = \{h_1, \ldots, h_l\}$, the base model of LSH iTables partitions the subsample into $2^l$ grid cells where the grid cell size is varied and determined by the random cut values of $h_i$.
Similarly, RS-H's base model builds a more fine-grained grid cell but its cell size is fixed and governed by the scaling $f$.
Therefore, LSH iTables add more diversity into the ensemble and hence leads to  higher average AUC scores.

\textbf{Compare with iForest.} LSH iTables shares similar average performance since RF-LSH is derived from the splitting mechanism of iForest.
We note that iForest uses RF-LSH whose weight, i.e. $max_i - min_i$, of dimension $i$ of each node is different from other nodes.
Since iForest uses the height limit of~8, iForest and LSH iTables share the same performance on many high dimensional data sets, e.g. $d > 20$, e.g. Musk, KDD99, Satimage-2, and Mnist.
On low dimensional data sets, including BreastW, Pima, Thyroid and Shuttle, the chance to select a dimension several times to build the tree is higher.
Hence, iForest and LSH iTables's performance are different.
LSH iTables is better on BreastW and Pima but worse on Thyroid and Shuttle.
Indeed, iForest can be seen as an LSH iTables variant that uses dependent hash functions, which make it hard to understand the behavior, especially to guarantee its detection accuracy.

\textbf{LSH iForest and ACE.} They show inferior average AUC scores.
ACE utilizes SimHash~\cite{SimHash} for the cosine similarity, which is not stable for detecting outliers.
ACE achieves highest AUC scores on Satimage-2, Pendigits, and Mnist, but suffers from worst AUC on many data sets, including Cardio, BreastW, Pima, and Cover with at most 54\%. 
LSH iForest utilizes RP-LSH for $\ell_2$ distance with a \emph{fixed} interval $w = 4$, and shares the same behavior with ACE.
LSH iForest shows inferior AUC scores on Cover, KDD99, Musk, and Mnist since the fixed $w$ limits the ensemble diversity. 

\textbf{LODA, Ensemble KNNW and Ensemble LOF.} They give similar average AUC scores and average ranking.
However, Ensemble KNNW and LOF run extremely slow while LODA's performance degrade significantly if using subsampling.
This is because LODA optimizes the bin size of its histogram based on the subsample.
Such learned values might be not optimal for the whole data.

Table~\ref{tb:Time} shows the running time of all studied ensemble methods where testing time dominates training time since $n \gg s$.
It is clear that LSH iTables is the fastest solution, followed by iForest and RS-H.
These methods runs in $\BO{mnl}$ time where $l$ is the number of hash functions used in hashing-based approaches and the height limit of iForest.
Since LSH iTables and iForest using the same hashing mechanism, they share similar running time.
We observe that the average number of hash functions used in LSH iTable is bounded by~8, which is setting of the height limit of iForest.
This explains why LSH iTables is slightly faster than iForest on Cover, the largest data set with~286,048 instances.
On Cover, RS-H is nearly 2 times slower than LSH iTables.
This is due to the extra cost of searching on CountMin sketches.

Ensemble KNNW and LOF run extremely slow with time complexity of $\BO{mnds}$.
LSH iForest and ACE share similar running time of $\BO{mndl}$ where $l$ is the number of hash functions of ACE and the height limit of LSH iForest.
The additional multiplicative $d$ on the complexity is due to the $\BO{d}$ RP-LSH evaluation.
LODA runs faster than ACE since it uses only~1 random projection. 
However, the computational overhead of learning the optimal bin size sometimes degrades the running time, as can be seen on KDD99 with more than 1 hour.




\subsection{LSH iTables on decentralized data}

\begin{figure*} [!t]
	\centering
	\includegraphics[width=1\textwidth]{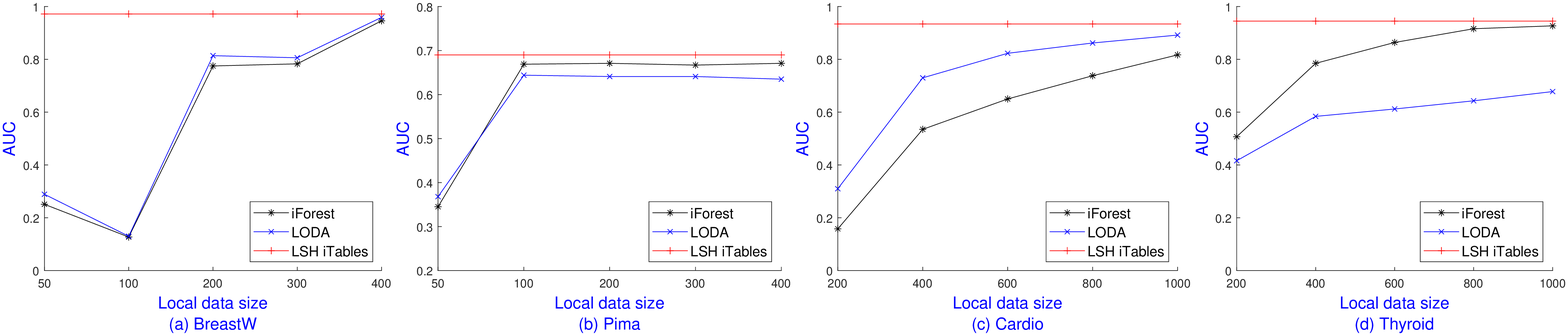}		
	\caption{AUC of iForest and LODA while varying the local data size.}
	\label{fig:Local}
\end{figure*}

We emphasize that iForest and LODA are not suitable for the decentralized scenario where participants release only local trained models on local data. 
Without the training data, they are not mergeable since iForest loses the isolation mechanism and LODA has different optimal bin sizes.
Indeed, decentralized iForest and LODA are just  ensemble methods where each base model are constructed on local data.
Therefore, their performance has deteriorated if one local data contains a majority of outliers.
Figure~\ref{fig:Local} show their accuracy when trained on various local data size with most of outliers on medical data sets.

It is clear that iForest and LODA's performance have deteriorated when trained on a limited subsamples with majority of outliers.
Their performance significantly increases when the local model has enough data.
Due to the mergeablity, LSH iTables has stable performance and each participant can form the global LSH iTables from local LSH iTables to achieve high detection rates.

\begin{figure*} [!t]
	\centering
	\includegraphics[width=1\textwidth]{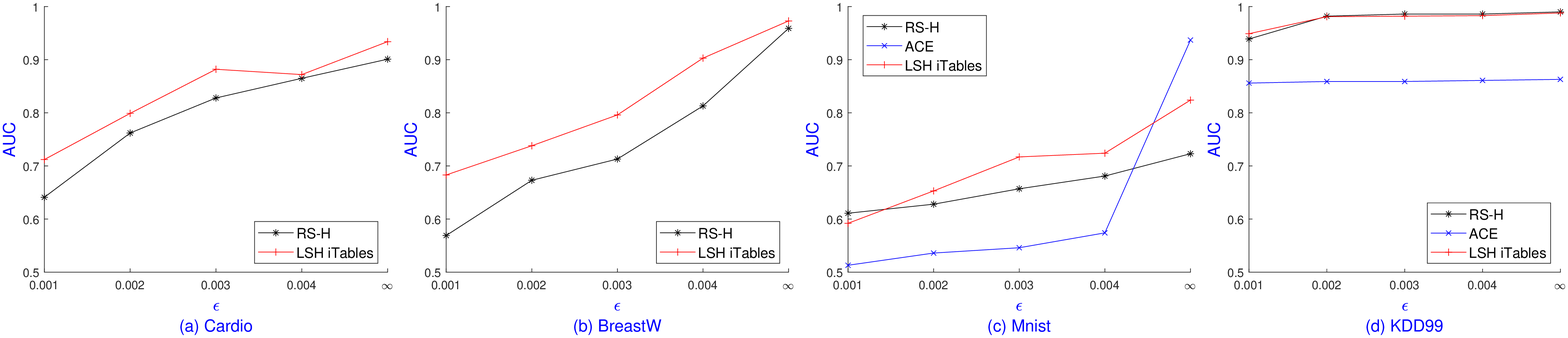}		
	\caption{AUC of RS-H, LSH iTables, and ACE under $\epsilon$-DP over a wide range of $\epsilon$.}
	\label{fig:DP}
\end{figure*}

We now evaluate the hashing-based approaches, including LSH iTables, RS-H and ACE given released $\epsilon$-DP histograms to detect collaborative outliers where multiple participants contain a part of the data set.

We consider two participants who share their local model to each other and report the AUC as the average AUC of two participants.
Each participant uses $\epsilon = \{0.001, 0.002, 0.003, 0.004\}$ and add Lap$(1/\epsilon)$ noise into the local histogram.
Figure~\ref{fig:DP} shows the AUC scores over a wide range of $\epsilon$ on hashing-based approaches and $\epsilon = \infty$ is the setting of non-privacy.
Note that we do not report ACE's performance on Cardio and BreastW due to its inferiority.

All methods provide higher accuracy when increasing the privacy budget $\epsilon$ and finally reach the non-privacy level.
ACE is very sensitive to $\epsilon$ on Mnist since its $\epsilon$-DP version gives the lowest AUC score whereas its non-privacy version gives the highest score.  
This is due to the fact that both LSH iTables and RS-H use the $\log$ scale of the bucket size as outlier score, which reduces the impact of Laplacian noise.
$\epsilon$-DP LSH iTables achieves higher AUC scores than RS-H, and the gaps are significantly larger than that of non-privacy version. 
This is because RS-H uses CountMin sketches with more added Laplacian noise than LSH iTables.
This observation has been shown again on Table~\ref{tb:Parties} where we fix $\epsilon = 0.01$ and vary the number of participants from 2 to 10.

\begin{table*}[b]
	\centering
	\caption{Comparison of AUC of $\epsilon$-DP LSH iTables and RS-H with $\epsilon = 0.01$.}
	\begin{tabular}{|c|c|c|c|c|c|c|c|c|c|c|}
		\hline
		& \multicolumn{5}{c|}{BreastW}            & \multicolumn{5}{c|}{Cardio}  \\  \cline{1-11}
		\# participants & 2 & 4 & 6 & 8 & 10 &           2 & 4 & 6 & 8 & 10          \\ \hline
		RS-H & 95.1 & 91.6 & 87.9 & 81.0 & 74.9 & 90.1 & 88.9 & 89.0 & 86.6 & 85.3 \\ 
		\hline
		LSH iTables & \textbf{97.0} & \textbf{92.4} & \textbf{92.1} & \textbf{82.6} & \textbf{78.3} & \textbf{91.8} & \textbf{91.4} & \textbf{90.2} & \textbf{89.5} & \textbf{88.1} \\
		\hline
	\end{tabular}
	\label{tb:Parties}
\end{table*}

\section{Conclusion}

We study the collaborative outlier detection where multiple participants collaborate to detect outliers on their local devices by only sharing their local detectors due to the privacy concern.
We propose \emph{LSH iTables}, an LSH-based ensemble method, with the mergeablity and compatibility with $\epsilon$-DP mechanisms.
On centralized data, LSH iTables outperform many advanced ensemble method in terms of accuracy and efficiency.
On decentralized data, $\epsilon$-DP LSH iTables returns similar accuracy with the non-privacy version given small privacy loss budgets $\epsilon$. 

\bibliographystyle{plain}
\bibliography{references}

\end{document}